# Indexing medical images based on collaborative experts reports

**Abir MESSAOUDI**
Department of Computer Sciences
High Institute of Management
BESTMOD Laboratory
Tunis, Tunisia
abir.messaoudi@gmail.com

**Riadh BOUSLIMI**
Department of Computer Sciences
High Institute of Management
BESTMOD Laboratory
Tunis, Tunisia
bouslimi.riadh@gmail.com

**Jalel AKAICHI**
Department of Computer Sciences
High Institute of Management
BESTMOD Laboratory
Tunis, Tunisia
jalel.akaichi@isg.rnu.tn

## ABSTRACT
A patient is often willing to quickly get, from his physician, reliable analysis and concise explanation according to provided linked medical images. The fact of making choices individually by the patient's physician may lead to malpractices and consequently generates unforeseeable damages. The Institute of Medicine of the National Sciences Academy (IMNAS) in USA published a study estimating that up to 98,000 hospital deaths each year can be attributed to medical malpractice [1]. Moreover, physician, in charge of medical image analysis, might be unavailable at the right time, which may complicate the patient's state. The goal of this paper is to provide to physicians and patients, a social network that permits to foster cooperation and to overcome the problem of unavailability of doctors on site any time. Therefore, patients can submit their medical images to be diagnosed and commented by several experts instantly. Consequently, the need to process opinions and to extract information automatically from the proposed social network became a necessity due to the huge number of comments expressing specialist's reviews. For this reason, we propose a kind of comments' summary keywords-based method which extracts the major current terms and relevant words existing on physicians' annotations. The extracted keywords will present a new and robust method for image indexation. In fact, significant extracted terms will be used later to index images in order to facilitate their discovery for any appropriate use. To overcome this challenge, we propose our Terminology Extraction of Annotation (TEA) mixed approach which focuses on algorithms mainly based on statistical methods and on external semantic resources.

## 1. INTRODUCTION
Currently, the unprecedented growth of social network media in various fields together with the amazing propagation on the global scale of web 2.0 social applications, have provided strong innovation in the information technology and an interesting part of mainstream culture. However, the use of these applications in the health care domain is still in its infancy [2]; this is due the fact that personal health information is recognized as highly sensitive. Moreover, physicians are usually late to embrace new information technologies; however, there are already e-health, mobile and pervasive health services such those provided through web sites, portals, or mobiles devices equipped with positioning technologies.
Social networking applications pledge to do more than just providing access to personal health information; one of its paramount tasks is to promote collaboration between patients, caregivers and health care providers. In [3], the authors consider that the participation in the health care management can render patients longer health conscious. Though, the main objective behind these up-to-date medical networks is to foster collaboration between the various actors and to place the patient at the heart of the health system [4].
The goal of this paper is to provide to physicians and patients, a social network that permits to foster cooperation and to overcome the problem of unavailability of doctors on site any time. Therefore, patients can submit their medical images to be diagnosed and commented by several experts instantly. Consequently, the need to process opinions and to extract information automatically from the proposed social network became a necessity due to the huge number of comments expressing specialist's reviews. For this reason, we propose a kind of comments' summary keywords-based method which extracts the major current terms and relevant words existing on physicians' annotations. The extracted keywords will present a new and robust method for image indexation. In fact, significant extracted terms will be used later to index images in order to facilitate their discovery for any appropriate use. To overcome this challenge, we propose our TEA mixed approach which focuses on algorithms mainly based on statistical methods and on external semantic resources.
This paper is organized as follows

## 2. RADIOLOGIST SOCIAL NETWORK
Nowadays, progress in medical imaging in both the hardware and software, and the interpretation techniques represents a very fruitful area in terms of research and innovation. Adding to that, social networks, such as PatientsLikeMe, radiolopolis, docadoc, Carenity and Sermo, are now offering new platforms of collaboration and knowledge exchange through the huge number of medical cases including opinions of professionals expressed over and done with an enormous number of comments. Medical images and comments are among the major fields meant for interaction between patients and physicians. The collection of annotations and/or reports from different experts is useful to better serve patients and to overcome the problem of availability of physicians on examination sites.

The first goal of our work is to design a social network dedicated for radiologists.

According to [2], the basic model of a social network in health care should include a :

- Set of patients that store their personal information in their health profile.
- Set of providers, representing health care professionals that may interact with patients by annotating medical outputs, exchange messages with patients, etc.

- Set of mechanisms for exchanging information, such as message boards, groups, emails and profile posts.
- Set of relationship types characterizing patients and providers participation.

- Set of search functions by which users can locate information. The social network is, obviously, controlled by a site operator. The implemented social network addressed to radiologists contains all of these features indicated beyond. The figure 1 expresses the social network static structure and represents the focal points of the proposed model.

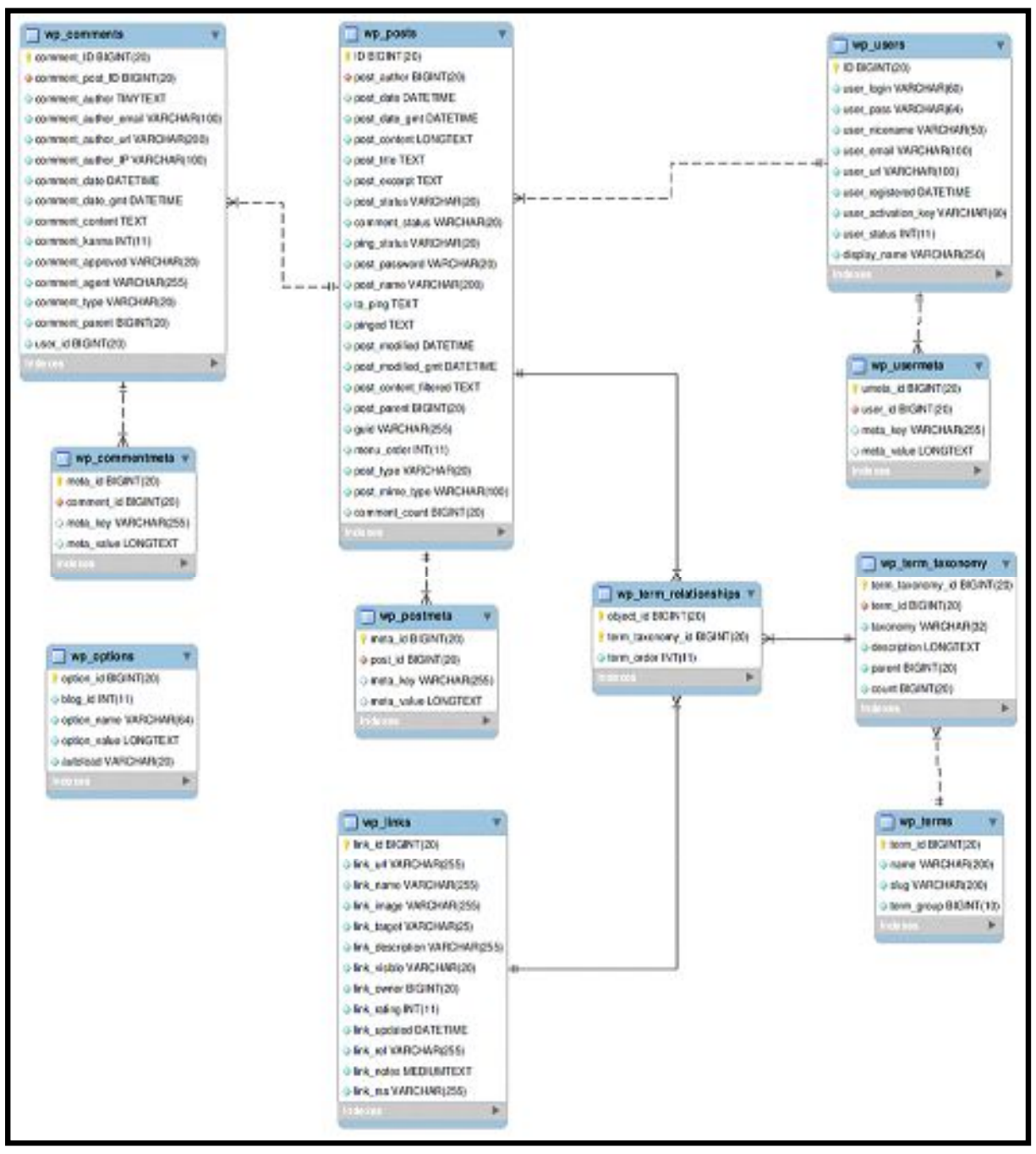

**Figure 1. Social Network Static Structure**

# 3. INDEXING MEDICALE REPORTS

The need to process automatically opinions and extract information, from the social network, is therefore becoming a necessity due to the huge number of comments expressing patients and physicians opinions. Information extraction aims to automatically putting out relevant information from texts for a particular task [8]. In our work, the paramount task is to extract powerful words from comments of physicians to index published medical images. Indexation is the process of describing and characterizing document using the representation of its contents. Its purpose is to indicate in a concise arrangement the content of the document, and to allow an efficient search of information in a collection of documents without having to analyze each of them for each user query [9]. In short, an index is a relation that connects each document to all keywords or descriptors describing the theme it treats. Descriptors can be single words, lemmas, terms of a thesaurus, external resource semantic concepts, summaries of paragraphs or any other unit of information that describe the contents of the document.

## 3.1 Indexation approaches

In broad literature, there exist three approaches of indexation: linguistic, statistical and mixed approaches. The linguistic approaches are based on a syntactic partial analysis or use patterns syntactic such as ontology, thesaurus, etc., to describe the content of the document. Several studies [10-13 have used Semantic Resources (RS) to perform the indexation task. For the statistical approaches, many research works have been proposed such as probabilistic latent semantic indexing [14] and latent Dirichlet allocation [15], etc. Several probabilistic or statistical combinations in the choice of weighting words were also used for indexation. According to [16-17], the frequency of occurrence of words in natural language texts is indicative of the importance of these words for the sole purpose of representing the content of the considered documents. Finally, the mixed approaches combine both syntax and statistics information to improve the accuracy in the detection of index terms. In [18], authors showed that the use of methods of extracting purely statistical terms provides results of equivalent quality, and then, it is not necessary to use linguistic tools adapted to a given language. Subsequently, they demonstrate that with an external semantic resource of sufficient quality, their purely statistical approach gives greater results to those using linguistic techniques. Thus, they find that with this statistical method, they will not have to change linguistic parser whenever document language changes. In addition, statistical approaches are simple to be implemented as opposed to linguistic ones.

## 3.2 The Terminology Extraction Of Annotation Approach

This section presents the proposed approach, which derives from mixed approach. It is a combination of statistical methods and an external semantic resource. The figure below represents the architecture of our Terminology Extraction Of annotation approach. (See the figure 2 below).

## 3.3 Detailed description

As illustrated, the Terminology Extraction Of annotation approach begins by the creation of the textual corpus containing comments on the medical image. The architecture includes six steps:

***Preprocessing:*** this step is composed of four preliminary steps: The decomposition of the corpus sentences, the removal of the punctuation points, the conversion of sentence to lowercase and the cutting of the sentence into words. In this step, we propose the pretreatment algorithm which contains four procedures, one for each sub-step .

***Cleaning:*** seeks to remove the *stopwords*. To achieve this, we use an anti-dictionary containing the most black words that seem useless in the medical field, we cite among them (many, how, again, which, since then, this, on some, but there, like why, however, when, which, soon...).

The anti-dictionary is a standard list contains *stopwords* not to be used as index such as prepositions, articles, pronouns, some adverbs and adjectives and some frequent words.

***Lemmatization :*** as comments may use different forms of a word for grammatical reasons, there exist families of derivationally related words. Lemmatization is used to regroup words in their belonging family. To find the lemmas, we implement the stemmer algorithm which seeks the root (prefix) and then assigns the suffix for a parent noun. Stemming algorithms are used in many types of language processing, text analysis systems, information retrieval and database search systems [19]. We use Porter Algorithm that is the best known algorithm for this work [20]. Porter's algorithm is too long and intricate ; it consists of 5 phases of word reductions, applied sequentially. Within each phase there are various conventions to select rules, such as selecting the rule from each rule group that applies to the longest suffix. We implemented the entire algorithm using all grammatical rules in French language.

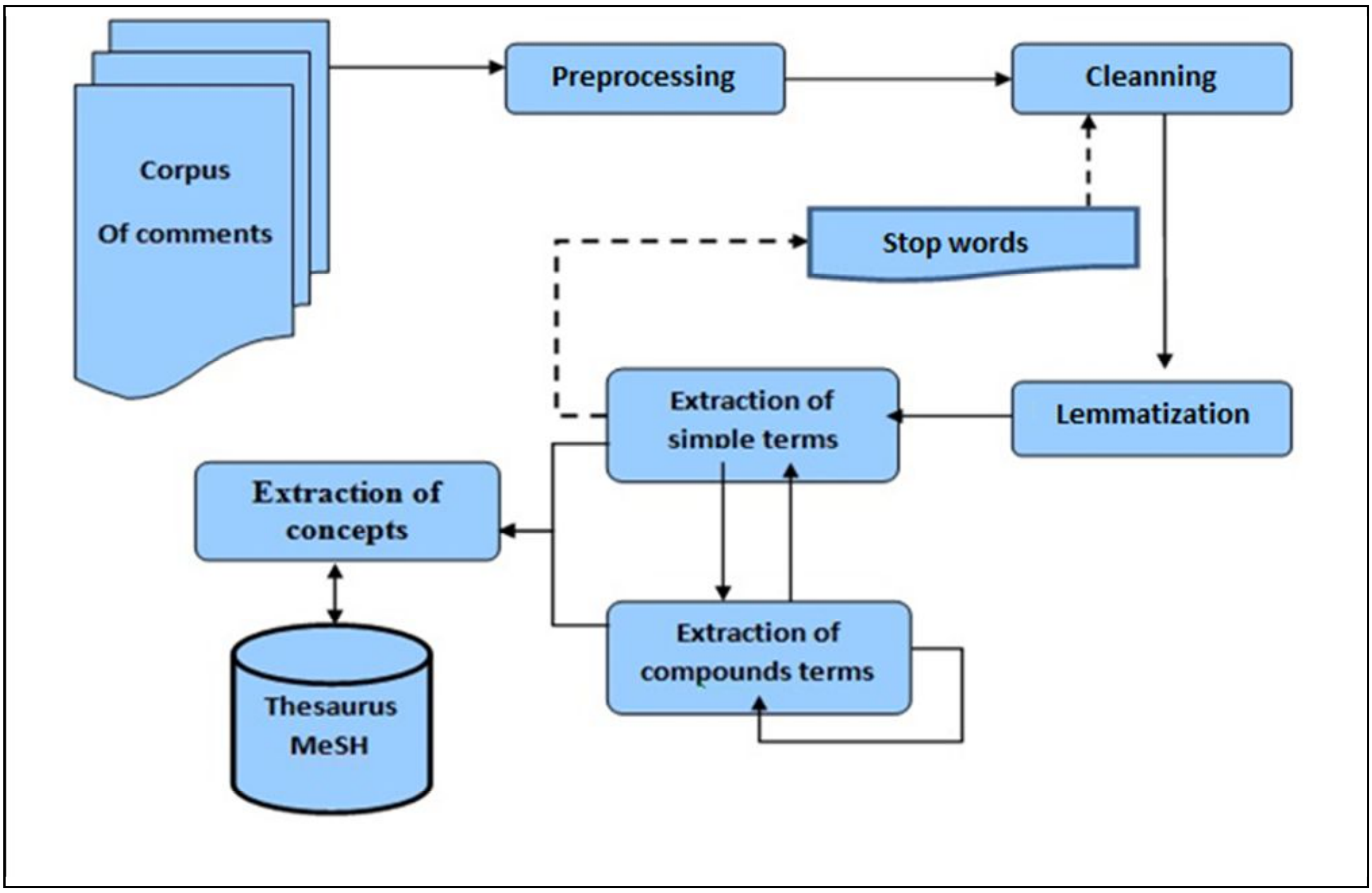

**Figure 2. Proposed model**

***Extraction of simple terms :*** in this milestone we measure the weight of each term using local and global weighting (tf.idf ) . This method combines two factors; the local weighting (TF) which quantifies the local representation of a term in the corpus of comments and the overall weighting (IDF) which measures the global representative of the term on the collection of corpus of comments.This measure rises with the number of occurrences within a document and with the scarcity of the term in the collection. tf.idf remains a best known weighting method in information retrieval (Kauchak, 2009). The formula is as follow:

$$\mathrm{tf.idf} = 0.4 + 0.6 * \frac{\mathrm{Tf}_{ij}}{\mathrm{Tf}_{ij+0.5+1.5\frac{\delta i}{\delta}}} * \frac{\log\left(\frac{N+0.5}{n_i}\right)}{\log(N+1)}$$

***Extraction of simple terms :*** in this milestone we measure the weight of each term using local and global weighting (tf.idf). This method combines two factors; the local weighting (TF) which quantifies the local representation of a term in the corpus of comments and the overall weighting (IDF) which measures the global representative of the term on the collection of corpus of comments.

- N: the total number of documents in the corpus ;
- $n_i$ : the number of documents containing term i ;
- $tf_{ij}$ :the local weighting of term i in document j ;
- $\delta_i$ : the length of document j in words ;
- $\delta$ : the average of the lengths of the documents of the corpus in words;

We keep only the terms that the output value exceeds a certain threshold set. In our case the (threshold =0.125).

***Extraction of compound terms:*** this step seeks to distinguish either it is a compound or a single word. Collocation measure attempts to check the terms which occur together more often, these terms often reduced to two or three words present a concept. This measure called mutual information (MI) given by CHURCH in 1990 compares the occurrence probability of co-occurrence words and the probability of these words separately. The process of extracting complex terms is an iterative and incremental process. We build complex terms of length n+1 words from the words of length n words. We start from the simple list of one word length, for each sequence of words we calculate the value of the MI. If the sequence of words exceeds the threshold set (threshold =0.15 in our case), the sequence will be comprised on the list of compound terms :

$$MI(m1, m2) = log\frac{P(m1, m2)}{P(m1) * P(m2)}$$

Where P(m1) and P(m2) are an estimation of the probability of occurrence of the words m1, m2 and P (m1, m2) is an estimation of the probability that the two words appear together.

***Extraction of concepts:*** this is a verification step which comes to use an external semantic resource, more precisely a medical thesaurus (MeSH) is used to filter keywords obtained in the previous step, to verify that the extracted term belong to the medical vocabulary. We propose the algorithm of Extraction of concepts.

Then, the general algorithm of comments indexation is:

**ALGORITHM IndexingComments**

```
INPUT
   d : Array [1..100] of File;
   d_i : File;
   t_i : String;
   TSimpleTerms      :  Array[1..9999] of String;
   TCompoundTerms :  Array[1..9999] of String;
OUTPUT
   TConcepts: Array[1..1000] of String
Begin
   For each d_i in d Do
   Begin
      While not EndOfFile(d_i) Do
      Begin
            Read(t_i)
            If StopWords(t_i) = True Then Cleaning(t_i);
      End;
    Lemmatization(ti,{t_i ∈ d_i});
    TSimpleTerms ← ExtractionofSimpleTerms({t_i ∈ d_i});
    TCompoundTerms ←
            ExtractionofCompoundTerms({t_i ∈ d_i},{t_i+1 ∈ d_i});
    TConcepts ← ExtractionofConcepts(TSimpleTerms,
                                       TCompoundTerms);
  End;
End.
```

## 4. EXPERIMENTAL RESULT

A Content Management System Word press was used in the development of our social network. Word press is currently the most popular blogging platform in use on the Internet as indicated **CMS Usage Statistics** in 2011. It has many features that allow creating and managing an entire website or just a blog. Codes were written in HTML, PHP, Ajax, and JavaScript, pages were designed using Dreamweaver and the data base used was MySQL. Some screenshots of the implemented radiologist social network are presented below in figure 3, 4, 5 and 6.

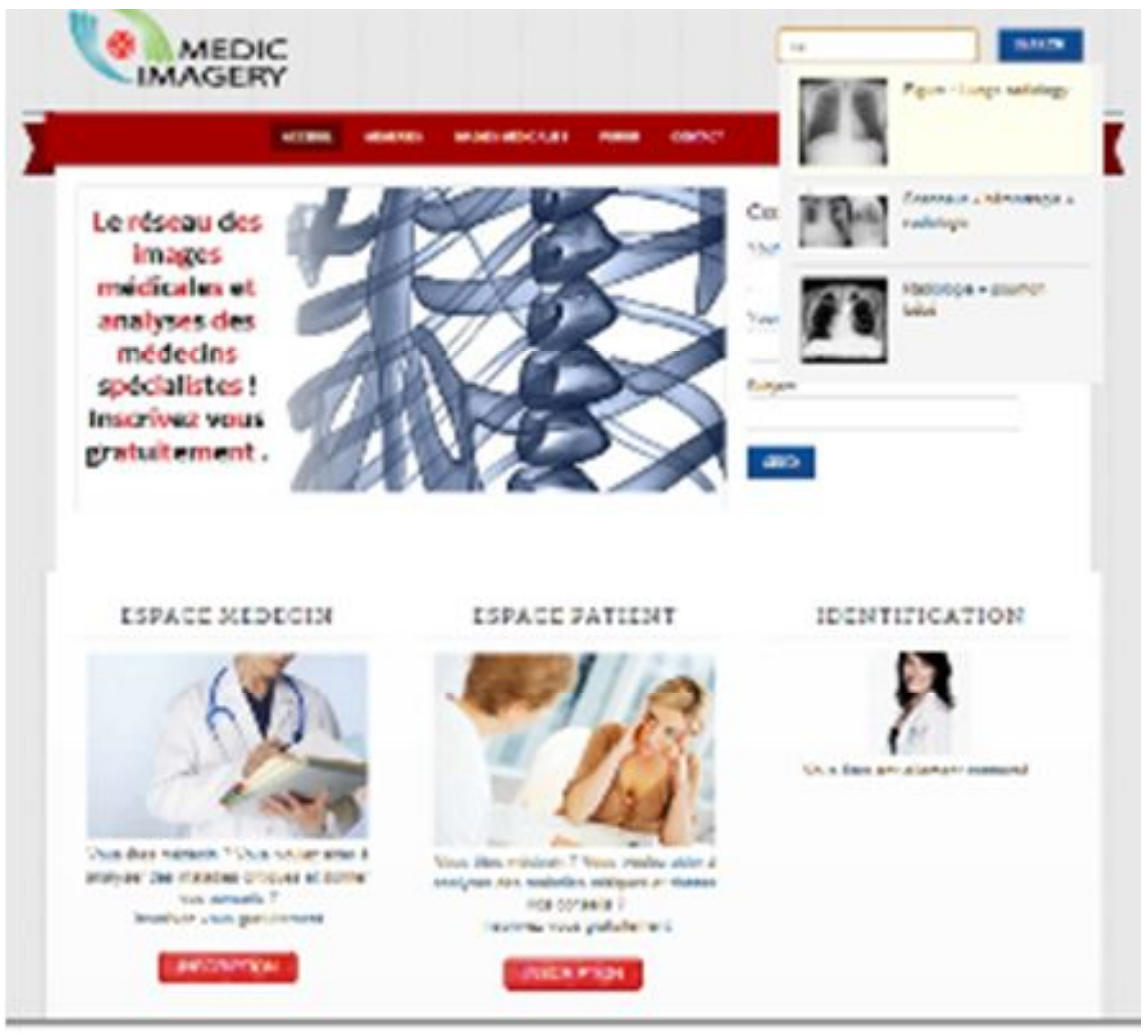

**Figure 3. Medic Imagery, the radiologist social network**

After developing the medical social network site, the challenge is to extract relevant information from comments to annotate and index images. Thirty examination cases from Tunisian physicians was obtained in various specialties for six different medical images. Each examination consists of a comment related on a medical image.

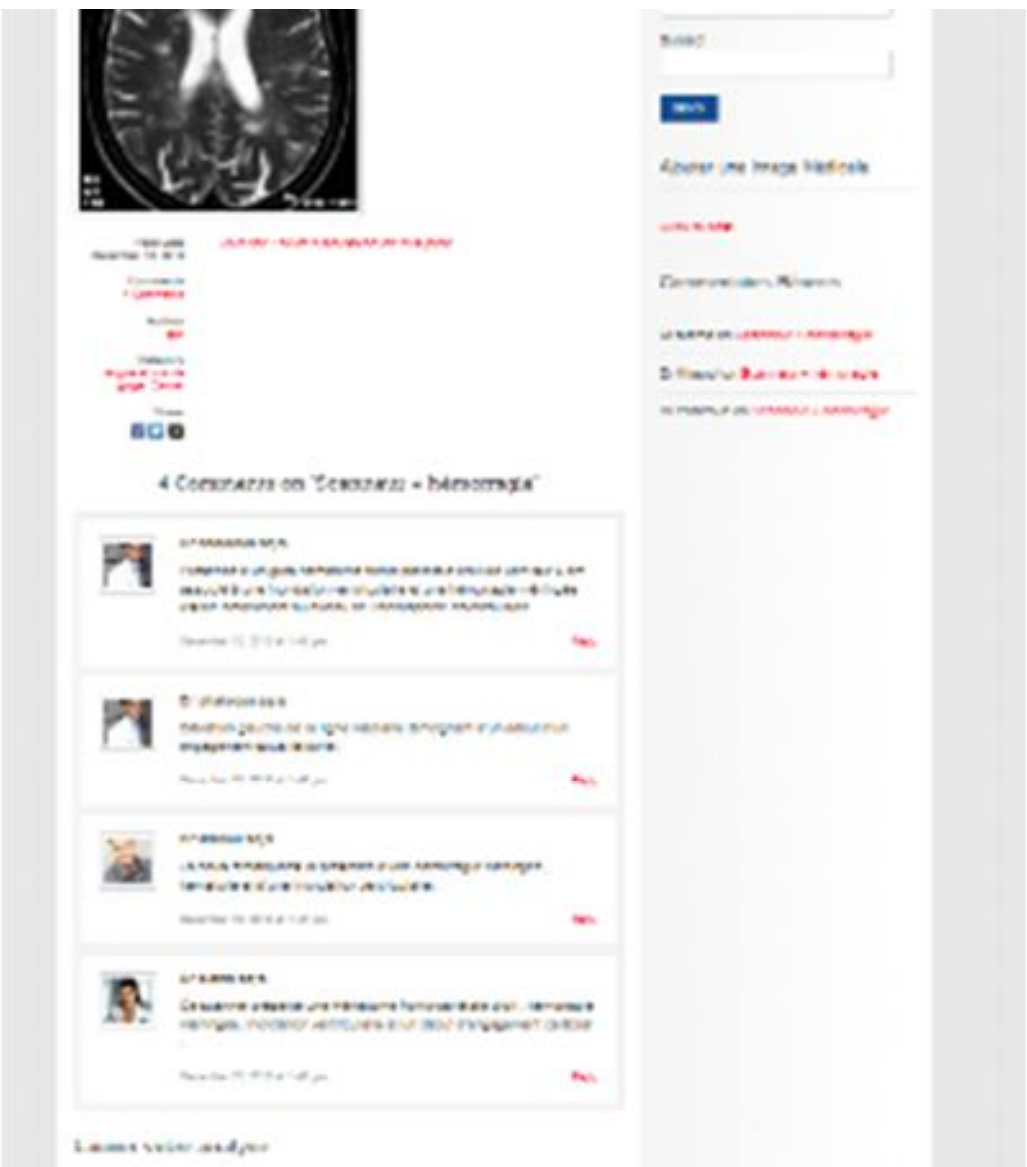

**Figure 4.Mixed posts from different physicians**

Methods described in section 3.2 were used to preprocess comments and to extract relevant medical terms those annotate and index images. Next section attempts to prove that the theoretical results are confirmed in practice.

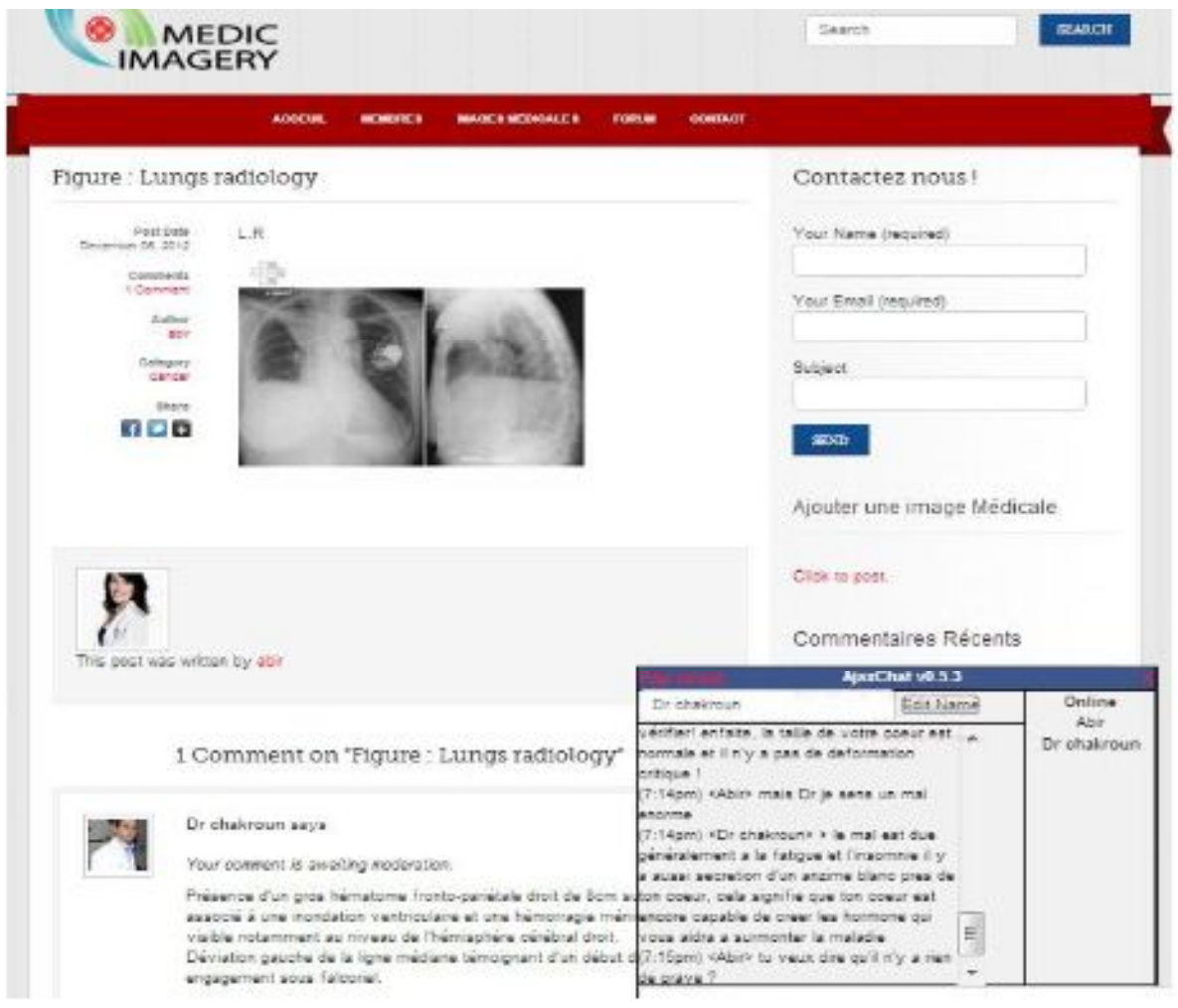

**Figure 5. Synchronous communication between a patient and a physician**

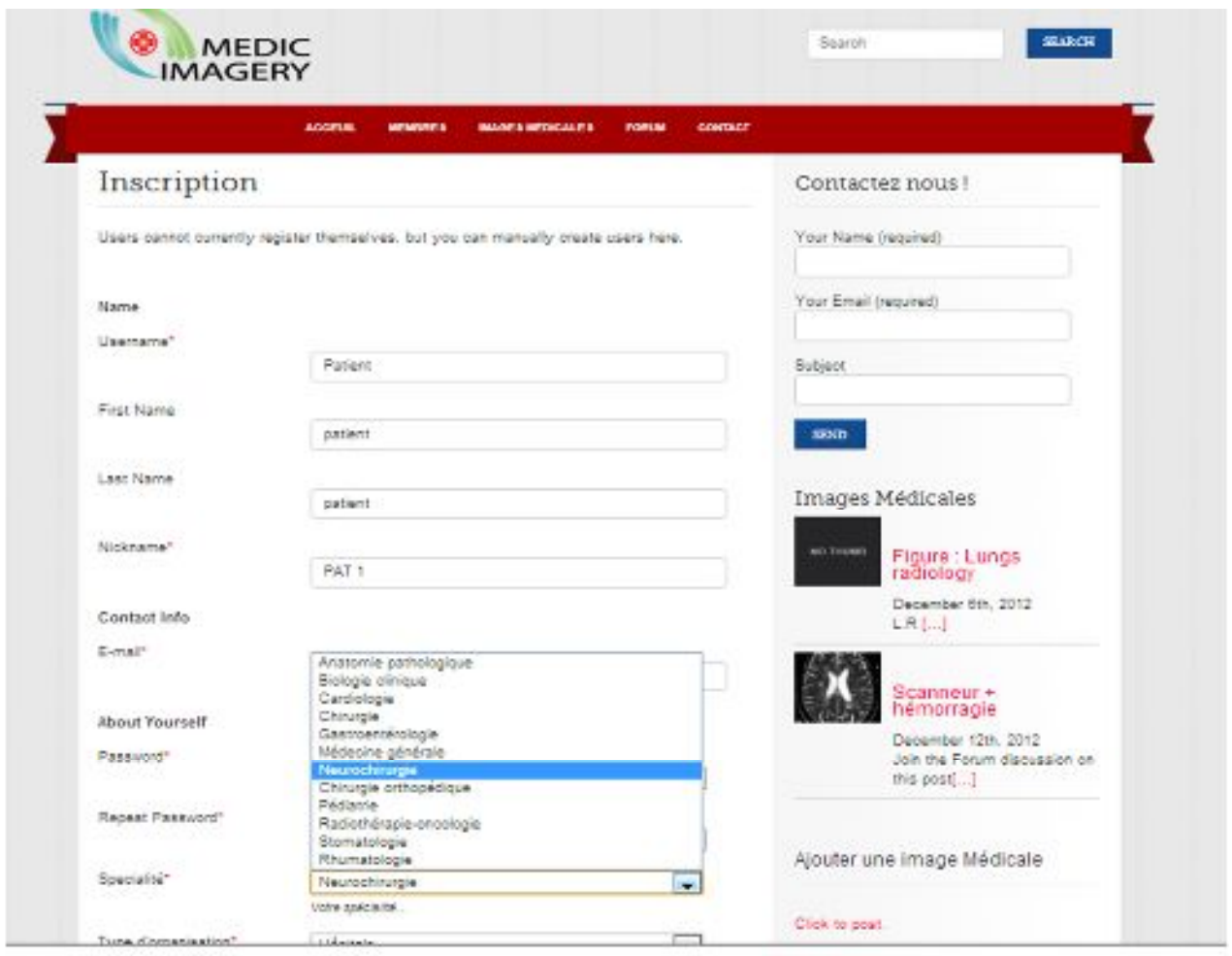

**Figure 6. Patient inscription**

Table 1 presents the full words extracted from the corpus of comments while applying the ***Preprocessing*** and ***cleaning*** steps.

**Table 1. Cleaning and Extraction of full words**

| Number | Annotation cleaned | Full words | |
|---|---|---|---|
| 1 | présence d un gros hématome fronto pariétale droit de 8 cm sur 4 cm associé à une inondation ventriculaire et une hémorragie méningée visible notamment au niveau de l hémisphère cérébral droit déviation gauche de la ligne médiane témoignant d un début d un engagement sous falcoriel | hématome pariétale inondation hémorragie hémisphère déviation engagement | fronto droit ventriculaire méningée cérébral gauche falcoriel |
| 2 | nous remarquons la présence d une hémorragie méningée hématurie et d une inondation ventriculaire | hémorragie hématurie ventriculaire | méningée Inondation |
| 3 | l image est peu nette pour décider mais nous constatons la présence d une hématome fronto pariétale piat et d une hémorragie méningée | image fronto piat méningée | hématome pariétale hémorragie |
| 4 | hémorragie méningée hématome début d engagement | hémorragie hématome | méningée engagement |
| 5 | cette radiographie présente une hématome fronto pariétale droit hémorragie méningée inondation ventriculaire et un début d engagement cérébral | radiographie fronto droit méningée ventriculaire cérébral | hématome pariétale hémorragie inondation engagement |
| 6 | c est une hématome fronto pariétale droit une hémorragie méningée | hématome pariétale hémorragie | fronto droit méningée |
| 7 | hémorragie méningée hématome fronto pariétal déviation de la ligne médiane | hémorragie hématome pariétal | méningée fronto déviation |
| 8 | il s agit d une hématome fronto pariétal encore une inondation ventriculaire et une hémorragie méningée | hématome pariétal ventriculaire méningée | fronto inondation hémorragie |

The table 2 shows the calculation of the TF.IDF .The threshold used for the tf.idf was set to 0.125 in our experiments. This value was subsequently adjusted after several experiments.

**Table 2. Calculate Tf-Idf**

| Full Words | Ann. 1 | Ann. 2 | Ann. 3 | Ann. 4 | Ann. 5 | Ann. 6 | Ann. 7 | Ann. 8 | AVG |
|---|---|---|---|---|---|---|---|---|---|
| **Hématome** | 0.0667 | 0.0000 | 0.1429 | 0.2500 | 0.0909 | 0.1667 | 0.1667 | 0.1429 | 0.1283 |
| **Fronto** | 0.0943 | 0.0000 | 0.2021 | 0.0000 | 0.1286 | 0.2358 | 0.2358 | 0.2021 | 0.1374 |
| **Parietal** | 0.0943 | 0.0000 | 0.2021 | 0.0000 | 0.1286 | 0.2358 | 0.2358 | 0.2021 | 0.1374 |
| **Droit** | 0.4000 | 0.0000 | 0.0000 | 0.0000 | 0.2727 | 0.5000 | 0.0000 | 0.0000 | 0.1466 |
| **Inondation** | 0.1333 | 0.4000 | 0.0000 | 0.0000 | 0.1818 | 0.0000 | 0.0000 | 0.2857 | 0.1251 |
| **Ventriculaire** | 0.1333 | 0.4000 | 0.0000 | 0.0000 | 0.1818 | 0.0000 | 0.0000 | 0.2857 | 0.1251 |
| **Hémorragie** | 0.0667 | 0.2000 | 0.1429 | 0.2500 | 0.0909 | 0.1667 | 0.1667 | 0.1429 | 0.1533 |
| **Méningée** | 0.0667 | 0.2000 | 0.1429 | 0.2500 | 0.0909 | 0.1667 | 0.1667 | 0.1429 | 0.1533 |
| **Hémisphére** | 0.2667 | 0.0000 | 0.0000 | 0.0000 | 0.0000 | 0.0000 | 0.0000 | 0.0000 | 0.0333 |
| **Cérébral** | 0.2667 | 0.0000 | 0.0000 | 0.0000 | 0.3636 | 0.0000 | 0.0000 | 0.0000 | 0.0788 |
| **Déviation** | 0.2667 | 0.0000 | 0.0000 | 0.0000 | 0.0000 | 0.0000 | 0.6667 | 0.0000 | 0.1167 |
| **Gauche** | 0.2667 | 0.0000 | 0.0000 | 0.0000 | 0.0000 | 0.0000 | 0.0000 | 0.0000 | 0.0333 |
| **Engagement** | 0.1610 | 0.0000 | 0.0000 | 0.6038 | 0.2195 | 0.0000 | 0.0000 | 0.0000 | 0.1230 |
| **Falcoriel** | 0.2667 | 0.0000 | 0.0000 | 0.0000 | 0.0000 | 0.0000 | 0.0000 | 0.0000 | 0.0333 |
| **Image** | 0.0000 | 0.0000 | 0.5714 | 0.0000 | 0.0000 | 0.0000 | 0.0000 | 0.0000 | 0.0714 |
| **Piot** | 0.0000 | 0.0000 | 0.5714 | 0.0000 | 0.0000 | 0.0000 | 0.0000 | 0.0000 | 0.0714 |
| **Radiographie** | 0.0000 | 0.0000 | 0.0000 | 0.0000 | 0.3636 | 0.0000 | 0.0000 | 0.0000 | 0.0455 |

After computing the simple terms whose the average exceeds 0.125, table 3 computes the mutual information as indicated in section III.2. The threshold used for the MI was set to 0.15 in our experiments. This value was subsequently adjusted after several experiments. The following table shows the calculation:

**Table 3 : Calculate Mutual information**

| Graphics | Hématome | Fronto | Pariétal | Droit | Inondation | Ventriculaire | Hémorragie | Méningée |
|---|---|---|---|---|---|---|---|---|
| **Hématome** | - | 0.234 | 0 | 0 | 0 | 0 | 0 | 0 |
| **Fronto** | 0 | - | 0.234 | 0 | 0 | 0 | 0 | 0 |
| **Pariétal** | 0 | 0 | - | 0.142 | 0.013 | 0 | 0 | 0 |
| **Droit** | 0 | 0 | 0 | - | 0.005 | 0 | 0.035 | 0 |
| **Inondation** | 0 | 0 | 0 | 0 | - | 0.178 | 0 | 0 |
| **Ventriculaire** | 0 | 0 | 0 | 0 | 0 | - | 0.035 | 0 |
| **Hémorragie** | 0 | 0 | 0 | 0 | 0 | 0 | - | 0.272 |
| **Méningée** | 0.07 | 0 | 0 | 0 | 0.035 | 0 | 0 | - |

The Compound terms extracted are:

- **Hématome fronto pariétale**
- **Hémorragie méningée.**
- **Inondation ventriculaire.**

**The last step** is the use of Mesh thesaurus for the extraction of concepts.

An excerpt of Mesh thesaurus is as follow :

```xml
<?xml version="1.0" encoding="utf-8" ?>
<rdf:RDF
  xmlns:rdf="http://www.w3.org/1999/02/22-rdf-syntax-ns#"
  xmlns:skos="http://www.w3.org/2004/02/skos/core#"
  xmlns:rdfs="http://www.w3.org/2000/01/rdf-schema#"
  xmlns:dc="http://purl.org/dc/elements/1.1/"
  xmlns:dcterms="http://purl.org/dc/terms/"
 xmlns:foaf="http://xmlns.com/foaf/0.1/">
  <skos:Concept rdf:about="http://www.fao.org/aos/agrovoc#c_1246">
    <skos:prefLabel xml:lang="fr">Cancer (genre)</skos:prefLabel>
    <skos:broader rdf:resource="http://www.fao.org/aos/agrovoc#c_2141"/>
    <skos:related rdf:resource="http!//www.fao.org/aos/agrovoc#c_1942"/>
  </skos:Concept>
 <rdf:Description rdf:about="&mesh;D000306"
    mesh:dateCreated="1999-01-01"
    mesh:dateEstablished="1979-01-01"
    mesh:dateRevised="2004-07-07"
    mesh:onlineNote="search ADRENAL GLAND NEOPLASMS 1966-74"
    mesh:publicMeSHNote="79,63-67; was see under ADRENAL GLAND NEOPLASMS 1975-78"
    skos:annotation="coord IM with histol type of neopl (IM)"
    skos:historyNote="Adrenal Gland Neoplasms (1966-1974)"
    skos:prefLabel="Adrenal Cortex Neoplasms"
    skos:scopeNote="Tumors or cancers of the ADRENAL CORTEX.">
    <mesh:activeMeSHYear>2005</mesh:activeMeSHYear>
  <mesh:activeMeSHYear>2006</mesh:activeMeSHYear>
  <skos:altLabel>Adrenal Cortex Cancer</skos:altLabel>
  <skos:altLabel>Cancer of Adrenal Cortex</skos:altLabel>
  <skos:altLabel>Cancer of the Adrenal Cortex</skos:altLabel>
  <skos:altLabel>Neoplasms, Adrenal Cortex</skos:altLabel>
  <skos:hiddenLabel>Adrenal Cortex Cancers</skos:hiddenLabel>
  <skos:hiddenLabel>Adrenal Cortex Neoplasm</skos:hiddenLabel>
  <skos:hiddenLabel>Adrenocortical Cancers</skos:hiddenLabel>
  <skos:hiddenLabel>Cancer, Adrenal Cortex</skos:hiddenLabel>
  <skos:hiddenLabel>Cancer, Adrenocortical</skos:hiddenLabel>
  <skos:hiddenLabel>Cancers, Adrenal Cortex</skos:hiddenLabel>
  <skos:hiddenLabel>Cancers, Adrenocortical</skos:hiddenLabel>
```

**Figure 7. An excerpt of Mesh thesaurus**

Finally, the extracted terms those index the brain radiation image loaded by a patient and annotated by several specialists are :

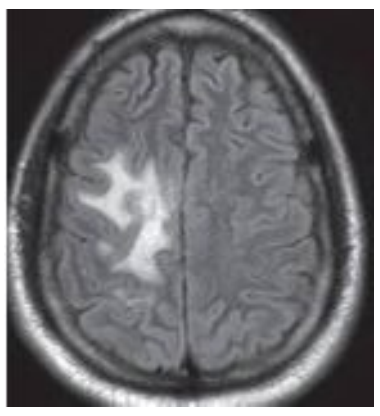

**Index Keywords** : Hématome fronto pariétale.
Hémorragie méningée.
Inondation ventriculaire

Similarly, our Terminology Extraction Of Annotation (TEOA) approach was used for this five different examination cases uploaded by patients in our **Medic Imagery** social network and reported by doctors and the extracted keywords which will be used later to index images was as in figure 8 follow .

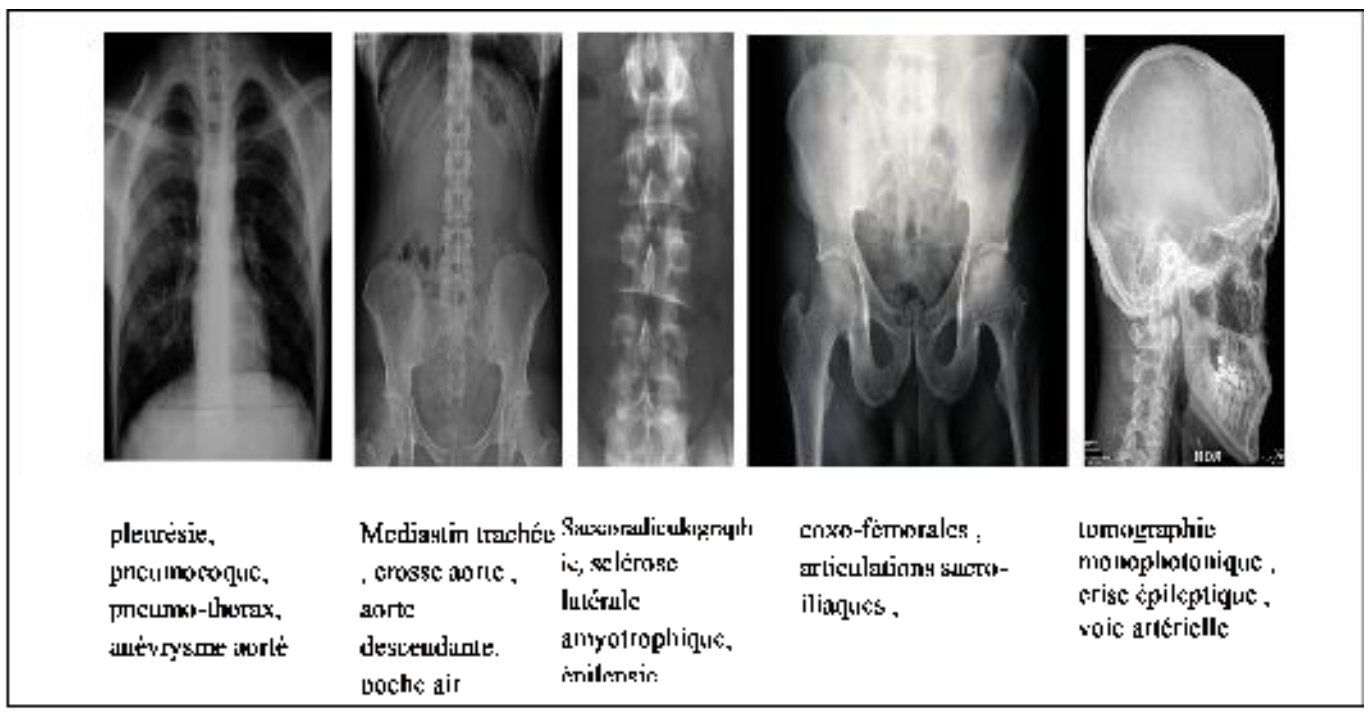

**Figure 8 : Extracted keywords from experts reports**

## 5. RESULTS

A preliminary step of our experiment is to index the medical reports using the proposed model that combines statistical methods and an external medical semantic resource. We evaluate the performance of the proposed method in the table 4 which summarizes the obtained values for each MAP testing. We note that our approach works well on extraction of relevant keywords for images compared to other systems [20-21]. These global observations are confirmed by the curves of precision / recall shown in Figure 8 . Figure 8 shows the difference in behavior between the three approaches.

| Type of approaches | Methods | MAP |
|---|---|---|
| **Statistical methods** | ANA | 0.1078 |
| **Linguistic methods** | Unitex | 1.1457 |
| **Mixed methods** | ACABIT | 3.2154 |
| | Xtract | 3.1745 |
| | Our system | 3.2450 |

**Table 1. Average precision results obtained for different types of radiographic images**

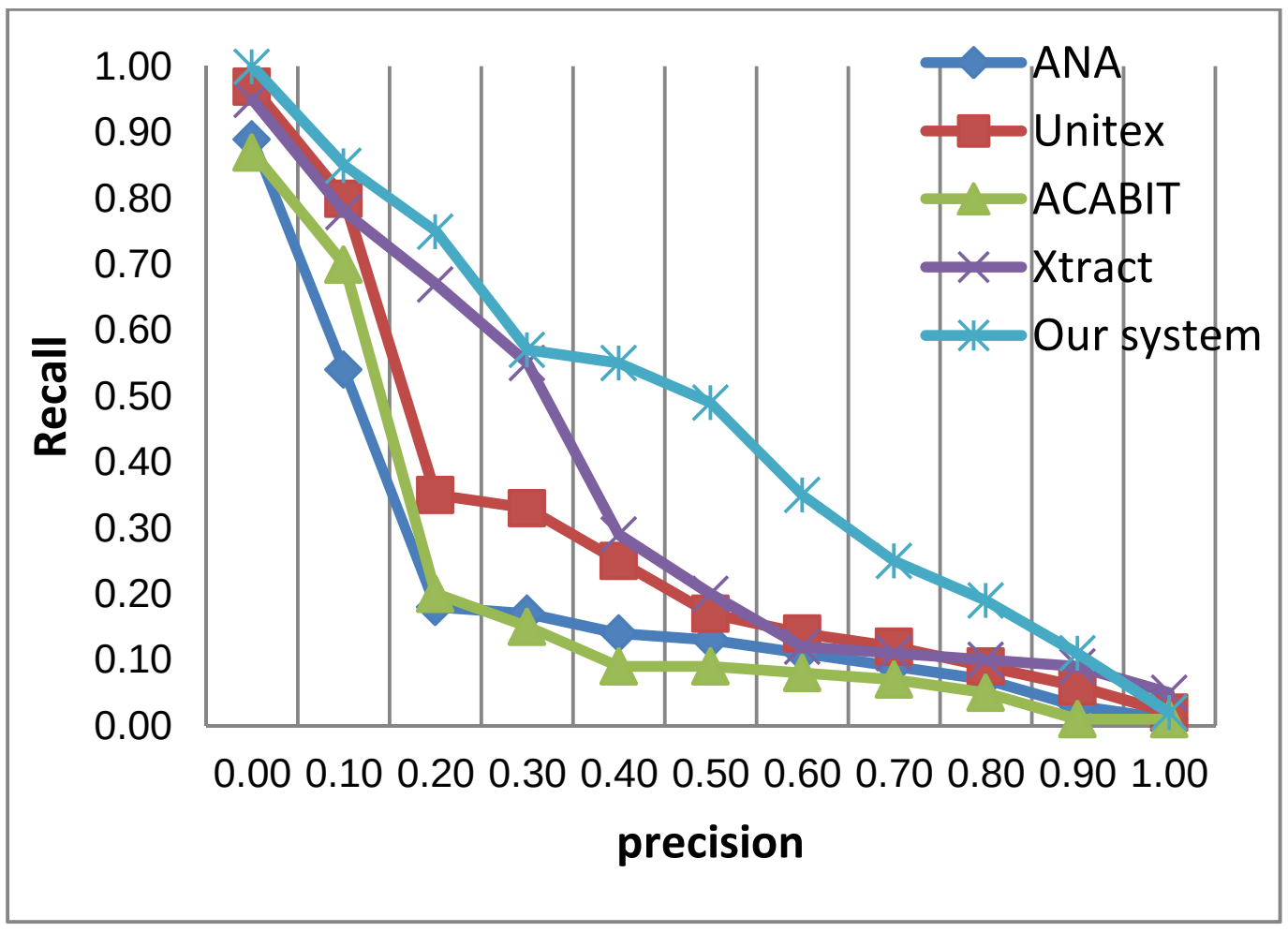

**Figure 8 : Courbe precision / recall for different Methods of radiological images indexation**

## 6. CONCLUSION

Place Given to the sensitivity of personal health information, we have been able to develop a medical social networking. This application will majorly serve to support collaboration and interaction. As being the decision maker and taker responsible for making choices on his own, the doctors may commit malpractices and generate unpredictable damage, now our medical social network place patient at the heart of the health system. Medical images and comments are among the major fields meant for interaction between patients and doctors, especially the radiologists. The collection of several annotations and reports from different experts is useful to better serve patients and to overcome the problem of availability of doctors on site. Therefore, the problem resides in the extensive data that formulate the comments and how to retrieve images after.

Further we propose our Terminology Extraction Of annotation (TEOA) approach that extracts hidden relevant words from reports to index images. This approach focuses on algorithm mainly based on statistical methods and an external semantic resource. Statistical methods used to select important and significant terms from comments. Then a verification step through a wealthy external medical semantic resource, more precisely a medical thesaurus (MeSH) is used to verify that those selected keywords belong to the medical field. This phase of extraction is done by the site administrator in front office.